\def\year{2019}\relax
\documentclass[letterpaper]{article} 

\pdfoutput=1
\usepackage{hyperref}

\usepackage{aaai19}  
\usepackage{times}  
\usepackage{helvet}  
\usepackage{courier}  
\usepackage{url}  
\usepackage{graphicx}  
\usepackage{float}
\usepackage{multirow}
\usepackage{stfloats}
\usepackage{url}
\usepackage{amsmath,amssymb,amsfonts} 
\usepackage{subfigure}
\usepackage{multirow}
\usepackage{epsfig}
\usepackage{graphicx}
\usepackage{latexsym} 
\usepackage{amsmath}
\usepackage{array}
\usepackage{float}
\newcolumntype{P}[1]{>{\centering\arraybackslash}p{#1}}
\newcolumntype{M}[1]{>{\centering\arraybackslash}m{#1}}

\graphicspath{{./images/}}

\begin{document}
%
\title{Future frame semantic segmentation of time-lapsed videos with large temporal displacement}
\author{ Talha Siddiqui \and Samarth Bharadwaj\\
IBM Research Labs, India\\
}
\maketitle

\begin{abstract}
An important aspect of video understanding is the ability to predict the evolution of its content in the future. This paper presents a future frame semantic segmentation technique to perform pixel-wise labeling of the current and future frames in a time-lapsed video. We specifically focus on time-lapsed videos with large temporal displacement to highlight the model's ability to capture large motions in time. 
We first introduce a unique semantic segmentation prediction dataset with over \emph{120,000} time-lapsed sky-video frames and all corresponding semantic masks captured over a span of five years in North America region. The dataset has immense practical value for cloud cover analysis, which are treated as non-rigid objects of interest. 
Next, our proposed recurrent network architecture departs from existing trend of using temporal convolutional networks (TCN) (or feed-forward networks), by explicitly learning an internal representations for the evolution of video content with time. Experimental evaluation shows an improvement of mean IoU over TCNs in the segmentation task by 10.8\% for 10 mins (21\% over 60 mins) ahead of time predictions. 
Further, our model simultaneously measures both the current and future solar irradiance from the same video frames with a normalized-MAE of 10.5\% over two years. These results indicate that recurrent memory networks with attention mechanism are able to capture complex advective and diffused flow characteristic of dense fluids even with sparse temporal sampling and are more suitable for future frame prediction tasks for longer duration videos. 

\end{abstract}

\section{Introduction}
To translate the significant progress made by the community in image semantic segmentation into better video understanding, memory networks are being employed to encode video content. Much like with humans, memory network allows systems to efficiently encode higher-order information content present in videos such as semantic structure across video frames. 
An interesting direction of research to enable deep neural networks in encoding this spatio-temporal semantic structure in videos is by future frame prediction \cite{neverova2017predicting,mathieu2015deep,srivastava2015unsupervised}. 
Pursing future frame prediction in videos allows us to build models that construct an internal representation of a video that captures the evolution of its content over time. Such a representation provides both pixel-level understanding that can be resolved to per-frame semantic segmentation, as well as frame-level property measurement, for both current and future frames. 

This paper generalizes the problem of video future frame semantic segmentation prediction to time-lapsed videos with large temporal time-steps. Further, a secondary task of measurement of frame level properties related to the content of the video is used as an additional constraint. Our framework jointly learns the dual objective and additionally localizes the pixel level contribution of the measurement implicitly. Lastly, we show empirical results for the specific application of prediction of cloud region from time-lapsed sky-videos and measurement of solar irradiance. 
\begin{figure}[!t]
\centering
\includegraphics[width=0.4\textwidth]{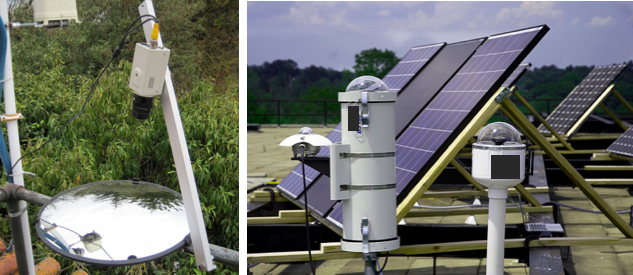}
\caption{\label{fig:fig1} Time-lapsed videos of meteorological phenomenon captured for recreational and scientific observations. Estimating the likely future weather patterns from such time-lapsed videos can influence project planning leading to social and economical benefits.}
\end{figure}

\subsection{Related Work}

Semantic segmentation techniques for an image, or pixel-wise labeling from structure, have made large strides beginning with the fully-convolutional augmentation of convolutional networks \cite{Long_2015_CVPR} and their variants with dilating or \emph{a'trous} convolutions \cite{yu2015multi}. In order to replicate similar success transitioning into video understanding, the computer vision community has embraced memory (recurrent) networks and more recently, temporal convolutional networks (TCN) \cite{tcn} in various interesting ways. 

We first position our work by briefly reviewing the recent flavours of memory networks to enhance video representations as follows: 

{\bf Encoding short video sequences} was initially viewed as the next step to extending the performance of image understanding to videos. Early approach used two tier architectures \cite{karpathy2014large} to compute features at different scales. Soon the encoder-decoder framework was favoured \cite{venugopalan2014translating} first by aggregation of frame representations and later with attention based approaches and 3D convolutional operations \cite{yao2015describing,tran2015learning}. \cite{klein2015dynamic} present a dynamic convolution approach to predict short term weather from radar imaging sequences. However, a sufficiently robust representation of video frames becomes challenging with increasing length and complexity in context. \cite{vondrick2015anticipating} approach anticipating the likely-labels in the future frames without flow.

{\bf Unsupervised next frame prediction} was introduced by \cite{srivastava2015unsupervised}, which provides video representations using LSTMs in the encoder-decoder framework. The reconstruction error is minimized in a composite encoder-decoder model that simultaneously predicts now and future frames. \cite{Shi:2015} extend the framework by introducing spatially constrained or convolutional LSTMs. The authors view the memory units as hidden layers and convert the input-state transition to convolution operations. These operations induce spatial structure in the memory units that is lacking in LSTM and can help encode evolving content in videos. Recently, \cite{JingyiHou_2018} utilize mid-level video frame encodings to capture semantic concepts in action recognition. 
 
{\bf Flow} estimation that is based on recurrent networks is implicitly computed and often intertwined with next frame prediction. Video encoder-decoder performance may improve with an additional penalty, added to the overall loss, of gradient smoothness parameters \cite{patraucean2015spatio}. Explicit optical flow approaches \cite{weinzaepfel2013deepflow,ilg2016flownet,feichtenhofer2016convolutional,hur2017mirrorflow} compute dense flow vectors for video understanding that track the motion of every pixel in the image. Such approaches are also shown to improve related tasks such as occlusion detection when exploited jointly. \cite{feichtenhofer2017detect} use the model architectures of explicit flow, two consecutive frames as simultaneous inputs, to improve both tracking and object detection in videos. 

{\bf Recurrent attention} is an aspect of semantic segmentation which identifies individual instances of class. While the approaches presented in literature are evaluated on images, they use memory networks for sequentially predicting objects. Graphical models have been replaced in favour of  recurrent networks with attention feedback, termed recurrent attention. \cite{romera2016recurrent} use a convolutional LSTM (ConvLSTM), while  \cite{ren2016end} later relax the spatial constraint for instance segmentation to capture disconnected instances of a given semantic class. Recently, \cite{piergiovanni2017learning} use temporal attention filters to identify latent sub-events in videos.  

{\bf Temporal convolutional networks} (TCN) \cite{tcn} refer to convolutional network architecture that are utilized for temporal prediction tasks such as next frame prediction. Unsupervised reconstruction of videos produces next frames but with noise and blurring, \cite{mathieu2015deep} replace the squared error loss with adversarial training method and show improvement in reconstruction. \cite{luc2016semantic} first propose using adversarial training for semantic segmentation. They next extend the work \cite{neverova2017predicting} to propose future frame and semantic segmentation prediction in the context automatic driving problem. Their approach uses convolutional filters with auto-regression for very short term forecasting of future semantic masks. While the approach has shown excellent results on the Cityscapes dataset \cite{Cityscapes}, the prediction task in the dataset is limited to 0.5 seconds future prediction. This may have limited practical use in automated driving applications. \cite{jin2017iccv} propose video scene parsing by using predictive feature learning and prediction steering parsing. Their predictive feature learning architecture learns to extract spatiotemporal features by enforcing the model to predict a future frame in the sequence. They extend their work \cite{jin2017nips} and propose simultaneous scene parsing and optical flow for future video frames. They state that capturing motion dynamics as well as predicting semantic masks are correlated problems and benefit from each other. However to the best of our knowledge, there is no recent work which illustrates the performance of spatiotemporal memory networks aided with spatial attention in predicting future frame semantic masks. Recently, \cite{miller2018recurrent} argue that the comparable performance of recurrent networks can often be achieved simply with feed-forward networks in an auto-regression fashion. Here, we show the merit of memory networks over auto-regression in time-lapsed videos with large temporal displacement.

\begin{figure*}[!ht]
\centering
\includegraphics[width=.9\textwidth]{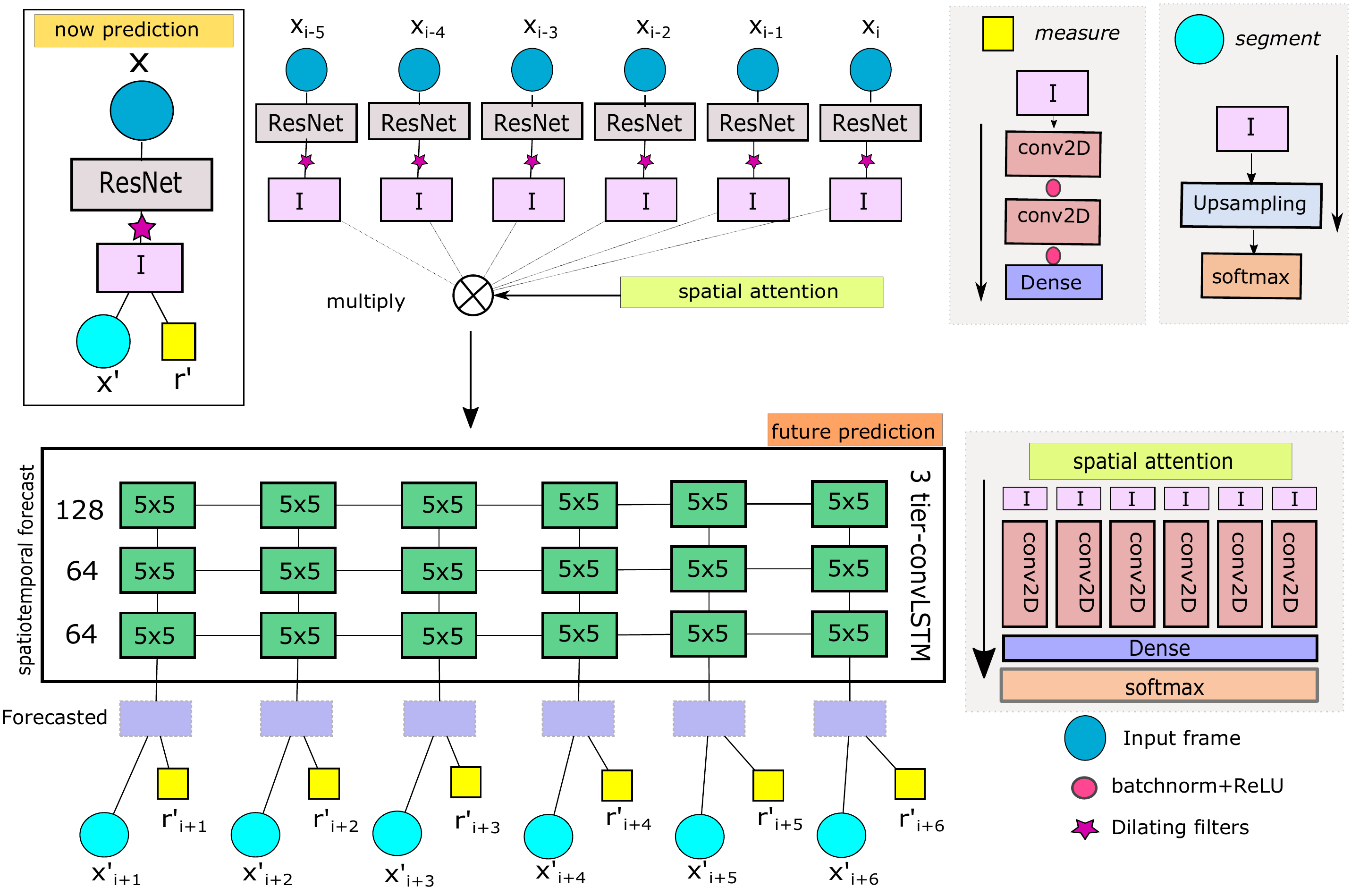}
\caption{[Best viewed in colour] The proposed \emph{future} video frame semantic segmentation approach explained here. For each videoframe ($x$), a representation tensor ($I$) is obtained using ResNet with dilating filters. $I$ is used in the \emph{now} model to predict both the current semantic mask ($x'$), with upsampling, and irradiance ($r'$), with convolutions, batchnorm+ReLU. The \emph{future} model is a 3-tier ConvLSTM over the frame representations $I$ of the previous $n$ frames. Further, spatial attention is applied on the frame representations to further enhance performance. (Details in Section Model).}
\label{fig:prop}
\end{figure*}

\subsection{Key Contributions}
The key contributions of this research can be summarized as follows:

\begin{itemize}
\item 
We propose a novel future semantic mask prediction framework for simultaneously performing two collegial tasks, namely \emph{segment} and \emph{measure} that often occur together in various applications of video understanding.

\item
We propose a memory network based approach for future frame segmentation in time-lapsed videos of weather phenomenon. We show that the performance of memory networks can be considerably amplified with spatial attention models and a dual objective.

\item
The time-lapsed Sky-video dataset introduced in this paper represents, in our view, a stronger challenge to the semantic segmentation prediction problem by relaxing two assumptions, object rigidity and temporal continuity. The dataset contains clouds at a frame rate of 6 frames per hour. 
    
\item
Our proposed approach out-performs state-of-art semantic pixel-wise labeling approaches that are designed for short-term videos on the sky-video dataset. The result makes a case for ConvLSTMs with attention as a middle ground between unsupervised recurrent nets and temporal convolutional feed-forward networks. 
\end{itemize}

\section{Model}
\label{sec:model}
We present a \emph{future} video-frame semantic segmentation approach designed for time-lapse videos of weather phenomenon. The proposed end-to-end trainable approach performs two tasks: \emph{segment} provides a semantic segmentation (or pixel-wise labeling) of the objects of interest in a current or future frames, and, \emph{measure} regresses and also forecasts a frame-level property related to the content of the video. Problems in recent literature, such as object detection and semantic segmentation prediction in videos \cite{neverova2017predicting}, partial object counting \cite{segui2015learning,chattopadhyay2016counting}, and video question answering (\cite{zeng2017leveraging}) can be viewed within our framework. 

We first describe the architecture of the model followed by our utilization of the spatial attention models to enhance the future prediction. We further describe the co-dependency in the architecture of the \emph{segment} and \emph{measure} components and argue that our architecture choice enables a spatially constrained representation to encode \emph{partial-}contributions of the measured property from all localized regions of a given video frame. 

\subsection{Notations}
 
\noindent Notationally, given a video $V$$\in$$\mathbb{R}^{N\times W\times H\times 3}$ which contains $N$ frames, $V^{1:N}$ with frames $\{x_1, x_2, x_3, \dots, x_N\}$, we propose a framework to determine a model ($\mathcal{M^\theta}$ and $\mathcal{M'^\theta}_{c}$), that produces both frame-wise semantic segmentation of $c$ classes of interest and their corresponding measurements, given by $\hat{V}^{1:t}_{\mathcal{M^\theta}}$=$\{(x'_1, r'_1), (x'_2, r'_2),$ $(x'_3, r'_3), \dots, (x'_t, r'_t)\}$, where $x'\in\mathbb{R}^{W \times H\times c}$, and  $r'\in\mathbb{R}$ is a scalar predicted value per frame. The \emph{now} and \emph{future} predictions for a look-back period $t$ are respectively obtained as follows: 

\begin{equation}
\begin{aligned}
(x', r')=\mathcal{M^{\theta}}(x) \\
 \hat{V}^{i:i+t} =\mathcal{M'^\theta}_{c} (V^{i-t:i})
 \end{aligned}
\end{equation}

\noindent The \emph{future} model ($\mathcal{M'^\theta}_{c}$) is built over an intermediary representation of an input video frame $I_i$ obtained from the \emph{now} model $\mathcal{M^\theta}$. Further, the model is learnt over a training set for $V$ with ground truth semantic segmentation $S$$\in$$\mathbb{R}^{N\times W\times H\times c}$, containing $\{y_1, y_2, y_3, \dots, y_N\}$, and the scale property measurement given by $\{m_1, m_2, m_3, \dots, m_N\}$.

\subsection{Architecture}
An overview of the proposed architecture for \emph{segment} and \emph{measure} for both \emph{now} and \emph{future} video frames prediction is illustrated in Fig. \ref{fig:prop}. We describe important details of the architecture that are relevant to the analysis, deferring remaining details of the model, such as parameters of layers, to the Supplementary. 

\noindent {\bf Now model}: Our approach utilizes the ResNet50 \cite{he2016deep} as the front-end model. We further augment the last convolutional layer with dilation \cite{yu2015multi}, to obtain the fully convolutional layer ($fc7$) that is used as a representation vector ($I$). Dilated convolutions are used to capture features at multi-scales without adding additional layers. All convolutional blocks are interlaced with Batch-norm and $ReLU$ non-linearity and use bi-linear up-sampling to merge the skip-connections.
 
The representational vector ($I$) obtained from the front-end model is bifurcated into the \emph{segment} branch, that is upsampled to the semantic segmentation mask resolution. Pixel label assignment is performed with \emph{softmax}. Conversely, the \emph{measure} branch is up-sampled with convolution up-sampling layers, again interlaced with Batch-norm and $ReLU$ non-linearity. The prediction measure is aggregated over a dense connection, with dropout, to a single value. 

\noindent {\bf Future model}: To perform future frame prediction, we use a recurrent architecture to predict tensor $I_i$ by utilizing only the corresponding representation tensors.\footnote{we use the term tensors rather than vectors to indicate their multi-dimensional nature.} ($\{I_{i-t}, \dots, I_i\}$) from historical frames. In order to preserve the \emph{spatial correspondence} of the $fc7$ representation, we utilize convolutional LSTMs (ConvLSTM) \cite{Shi:2015} that contain memory gates which are convolutional operations and preserve the spatial and temporal structure while inducing memory into the architecture. 

Specifically, multiple tiers of ConvLSTM are used over a look-back period $t$, to construct the representations ($\{I_i, \dots, I_{i+t}\}$). We also observe that the performance of the stacked ConvLSTMs is significantly improved with the addition of spatial attention mechanisms, described next. 

\subsection{Future prediction with spatial \emph{soft-}Attention}
Our architecture uses stacked convolutional LSTMs to predict semantic representations of the future frames, from a look back period $t$, given by $\mathbb{I}$=$\{I_{i-t},\dots, I_{i}\}$. Attention is computed over the spatial, temporal and class label ($c$) dimensions of $I$ to induce more structure in the predictions, that can particularly be affected by the large temporal displacements in time-lapsed videos. 

\noindent {\bf Spatial attention} is assigned to a given pixel $(i,j)$ of $\mathbb{I}$ over all the values in the look back period $t$, of the given and neighbourhood pixels ($p$), with a simple convolutional operation. This can be described as: 
\begin{equation}
A_{(i,j)} = \mathbb{I}(i,j) \circledast h(p,p)
\end{equation} 
\noindent where $h$ is an additional set of convolutional filters learnt in the overall architecture.  The filters are oversampled and reduced with a densely connected operation to obtain the attention mask $A$. The mask obtained is further normalized with \emph{softmax} and multiplied with the original samples $\mathbb{I}$. This attention model utilizes learned convolutional operations over the \emph{entire} training batch matrix, learning attention without explicit temporal structure or memory. 
We also compute \emph{softmax} attention \cite{olah2016attention} over $\mathbb{I}$ averaged over all the pixels and use for baseline comparison, that we term \emph{mean} attention. 

\subsection{Loss}
As shown in the   Fig. \ref{fig:prop}, the model's end-to-end trainable architecture performs both semantic segmentation prediction and measure prediction for both \emph{now} and \emph{future} problem. The loss function is a weighted mixture of the loss from each task, given in Eq. \ref{eq:loss}:

\begin{equation}
\label{eq:loss}
L_{w}(x, {x'}, r, {r'}) = \lambda ( CL_{\gamma, \alpha}(x, {x'}) ) + (\mathcal{H}(r, {r'})) 
\end{equation}

\noindent where $CL_{\gamma, \alpha}$ is the combined loss given in Eq. \ref{eq:combloss}, and the $\lambda$ is the normalizing weight to equally favour predictions of segment and measure respectively.  

\begin{equation}
\label{eq:combloss}
CL_{\gamma}(x, {x'}) := FL_{\gamma, \alpha}(x, {x'}) + CE(x, {x'})
\end{equation}
\begin{equation}
\label{eq:focalloss}
FL_{\gamma, \alpha}(x, {x'}) := - \alpha  x^{\gamma} \log({x'}) + (1- x)^{\gamma} \log (1-{x'})
\end{equation}

\noindent The combined loss is a combination of focal loss ($FL_{\gamma, \alpha}$) \cite{lin2017focal} and categorical cross-entropy ($CE$) measured for the semantic segmentation over every pixel. The focal loss is given by Eq. \ref{eq:focalloss} with tuneable parameters set to control the order of magnitude ($\gamma=2$ and $\alpha=0.5$). The combined loss penalizes incorrect classification of pixel semantic label, with larger focus on harder predictions, such as cloud pixels. We observe small improvement in segmentation compared to pixel-wise categorical cross entropy. 

\begin{equation}
\label{eq:huberloss}
\mathcal{H}(r, \hat{r}) := \log( m * \cosh(r - \hat{r}))
\end{equation}

\noindent To compute prediction errors of the frame level measurement, in this case the irradiance forecast, smooth Huber loss is used (Eq. \ref{eq:huberloss}). We prefer the hyperbolic version for the smoothness, which is magnified at $m=100$. Empirically, irradiance prediction below $100$ $W/m^2$ are less important. 

\begin{figure}[!ht] 
\centering
\includegraphics[width=0.2\textwidth]{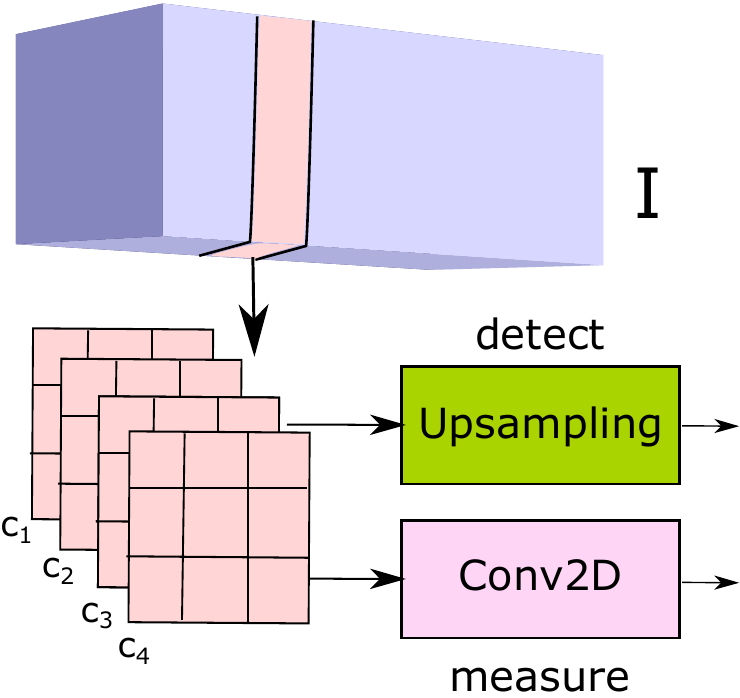}
\caption{\label{fig:zoom} An illustration of the bifurcation point of the model. Note $I$ is a tensor spatially consistent with the image rather than a flat vector.}
\end{figure}

\subsection{Segment with partial Measure}
\label{sec:partial}
A unique feature of our architecture is the strong constraint for every spatial region to have a \emph{partial}-contribution to the measured property $r$. As illustrated in Fig. \ref{fig:zoom}, the bifurcation point of the architecture that branches into two tasks, \emph{segment} and \emph{measure}, emerge from a single down-sampled representation $I$ (rather than a flat vector), which is received directly from the input image for \emph{now} prediction, and is predicted from the stacked convolutional-LSTMs for the \emph{future} prediction. In either case, for a given pixel, the \emph{segment} branch up samples $I$ using \emph{only} local neighbourhood vectors. Similarly, for the \emph{measure} branch, the convolution filters again operate only on the local neighbourhood of the pixel to predict $r$.  Extending this intuition to the \emph{future} prediction model, the stacked convolutional LSTMs are also constraint locally to predict the representation $I$ for a future frame. Hence, we assert that our framework implicitly generates the \emph{partial}-contribution of the measure property over the spatial region of the image. Hence, measure is an integral over a downsampled semantic mask prediction. 

\section{Evaluation}
In this section, we describe the performance of the proposed approach on newly introduced sky-video dataset. The proposed approach is compared on the same parameters and protocol with two different attention mechanisms and with a temporal convolutional network \cite{neverova2017predicting} designed for shorter videos. The baseline persistence model refers to a model which simply uses the predicted semantic mask of the current frame (at $t$) as the semantic mask of the future frame. 

\subsection{Sky-video dataset and protocol}
A sky-video is obtained from an upward facing wide-angle lensed video camera such as the one shown in   Fig. \ref{fig:fig1}. The dataset is recorded at Solar Radiation Research Laboratory (SRRL), Golden, Colorado \cite{srrl}\footnote{the dataset is  available for download at \url{www.nrel.gov/midc/srrl_bms/}. The processed dataset is available for easy reproducibility at \url{https://bit.ly/2Bw7HGP}.}, situated in North America. The time-lapsed videos are recorded using a commercial sky imager (TSI) \cite{morris2005total} at every 10 minutes interval. A mechanical sun tracker is used to block the sun preventing saturation in the image and blooming effects. The dataset is available for the last 13 years from 2005-2017. Over the same period, we obtained solar irradiance measurements (in $W/m^2$)from the same location using a pyranometer. The cloud cover, defined as the ratio of pixels labeled cloud in the sky image is correlated with irradiance measure (0.67 on the training set). More details of capture device and sample frames illustrations are in Supplementary.
\begin{table}[!ht] \small
\centering
		\caption{\label{tab:now}  normalized-MAE of Irradiance ($W/m^2$) (\%) and IoU for segmentation in Now prediction on the test set}
		\begin{tabular} {c c c c}
		\hline  
			Experiment & n-MAE  & IoU Cloud & IoU Sky \\\hline
			Baseline (only Irradiance) & 11.31 & - & - \\\hline
			ResNet50 + dilation & \bf 10.96 & \bf 83.31 & \bf 87.89\\\hline
		\end{tabular}
\end{table}
\noindent We showcase our experiments on the images captured in the year 2010-2014 with the total number of 123,064 images in our dataset. The input video consists of RGB frames of dimensions $W=320$, $H=320$. The TSI imager also provides ground truth segmentation masks of four classes in the sky, namely, \emph{sky, cloud, sun,} and \emph{tracker} ($c=4$). We use data from 2010-2012 in training (73,120 images) and from 2013-2014 (49,944 images) for testing based on availability and quality of ground truth. 

\noindent For the \emph{future} model, we use look-back ($t$)=$1$hour (or six samples at 10mins interval) and batch-size=$4$ ($3046$ batches per epoch) and compute a forward prediction for the same size. We start with a learning rate of $0.002$ and reduce with power decay  ($\gamma=0.9$) per epoch. A weight decay of 0.00005 with $l_2$-norm regularizer is used for all convolutional layers. \emph{Adam} optimizer is used for all our experiments. We report pixel-level segmentation accuracy, Intersection over Union (IoU) and normalized mean absolute error (nMAE) as applicable. 
\begin{table}[!ht] \small
	\centering
	\caption{\label{tab:measure} Measure Task: normalized-MAE (\%) of Irradiance (in $W/m^2$) for Future using proposed Attention Mechanisms on Testing Data}
	\begin{tabular}{c c c c} 
		\hline Attention Mechanism & +10 mins & +20 mins & +30 mins  \\\hline\hline
		Without & 28.057 & 28.472 & 28.229  \\\hline
		 \cite{olah2016attention} & 21.853 & 22.954 & 24.013 \\\hline
		Spatial & \bf 19.486 & \bf 22.051 & \bf 23.472 \\\hline
	\end{tabular}
	
	\begin{tabular}{c c c c} 
		\hline Attention Mechanism & +40 mins & +50 mins &+60 mins \\\hline\hline
		Without & 27.526 & 27.872 & 27.883 \\\hline
		 \cite{olah2016attention} & 25.048 & 26.210 & 27.173 \\\hline
		Spatial & \bf 24.934 & \bf 25.982 & \bf 26.938 \\\hline
	\end{tabular}
\end{table} 

\begin{table*}[!ht]
	\begin{center}
		\caption{\label{tab:seg} Segment Task: Accuracy(\%) and IoU for Future Frames using various Attention Mechanism on Testing Data}
		\begin{tabular}{ l  M{2cm} c c  c c  c c  }
		\hline
			 \multirow{1}{*}{Attention Mechanism} & \multicolumn{1}{c }{Accuracy}
			& \multicolumn{2}{c  }{+10 mins} & \multicolumn{2}{c  }{+20 mins} & \multicolumn{2}{c }{+30 mins} \\\cline{3-8}
			&  & Cloud & Sky & Cloud & Sky & Cloud & Sky\\\hline\hline
			Persistence & 80.55 & 64.04 & 69.77 & 59.31 & 65.65 & 55.97 & 62.74\\\hline
			Without & 87.59 & 61.61 & 69.76 & 61.36 & 69.57 & 61.15 & 69.55\\\hline
			Mean \cite{olah2016attention} & 89.00 &  70.39 & 77.25 & 68.39 & 75.34 & 66.15 & 73.38\\\hline
			Spatial & \bf 89.38 & \bf 74.15 & 79.57 &  70.48 &  76.43 & \bf 67.73 &  73.81\\\hline
			\cite{neverova2017predicting}& 75.94 & 63.15 & 68.88 & 58.08 &  64.08 & 52.64 & 56.72\\\hline
		\end{tabular}
		
		\begin{tabular}{ l  M{2cm} c c  c c  c c  }
		\hline
			 \multirow{1}{*}{Attention Mechanism} & \multicolumn{1}{c }{Accuracy}
			& \multicolumn{2}{c  }{+40 mins} & \multicolumn{2}{c  }{+50 mins} & \multicolumn{2}{c  }{+60 mins} \\\cline{3-8}
			&  & Cloud & Sky & Cloud & Sky & Cloud & Sky\\\hline\hline
			Persistence & 80.55 & 53.14 & 60.31 & 50.14 & 57.90 & 47.05 & 55.77 \\\hline
			Without & 87.59 & 60.97 & 69.49 & 59.71 & 69.11 & 58.66 & 69.53 \\\hline
			Mean \cite{olah2016attention} & 89.00 & 63.89 & 71.75 & 60.96 & 70.13 & 58.87 & 69.65 \\\hline
			Spatial & \bf 89.38 & 65.29 & 71.71 & 62.49 & 69.89 & \bf 60.16 & 69.05 \\\hline
			\cite{neverova2017predicting}& 75.94 & 48.02 & 49.61 & 43.76 & 42.89 & 39.55 & 35.53 \\\hline
		\end{tabular}
	\end{center}
\end{table*} 
\subsection{Analysis}
\label{sec:ana}
\begin{figure}[!t] 
\centering
\includegraphics[width=.45\textwidth]{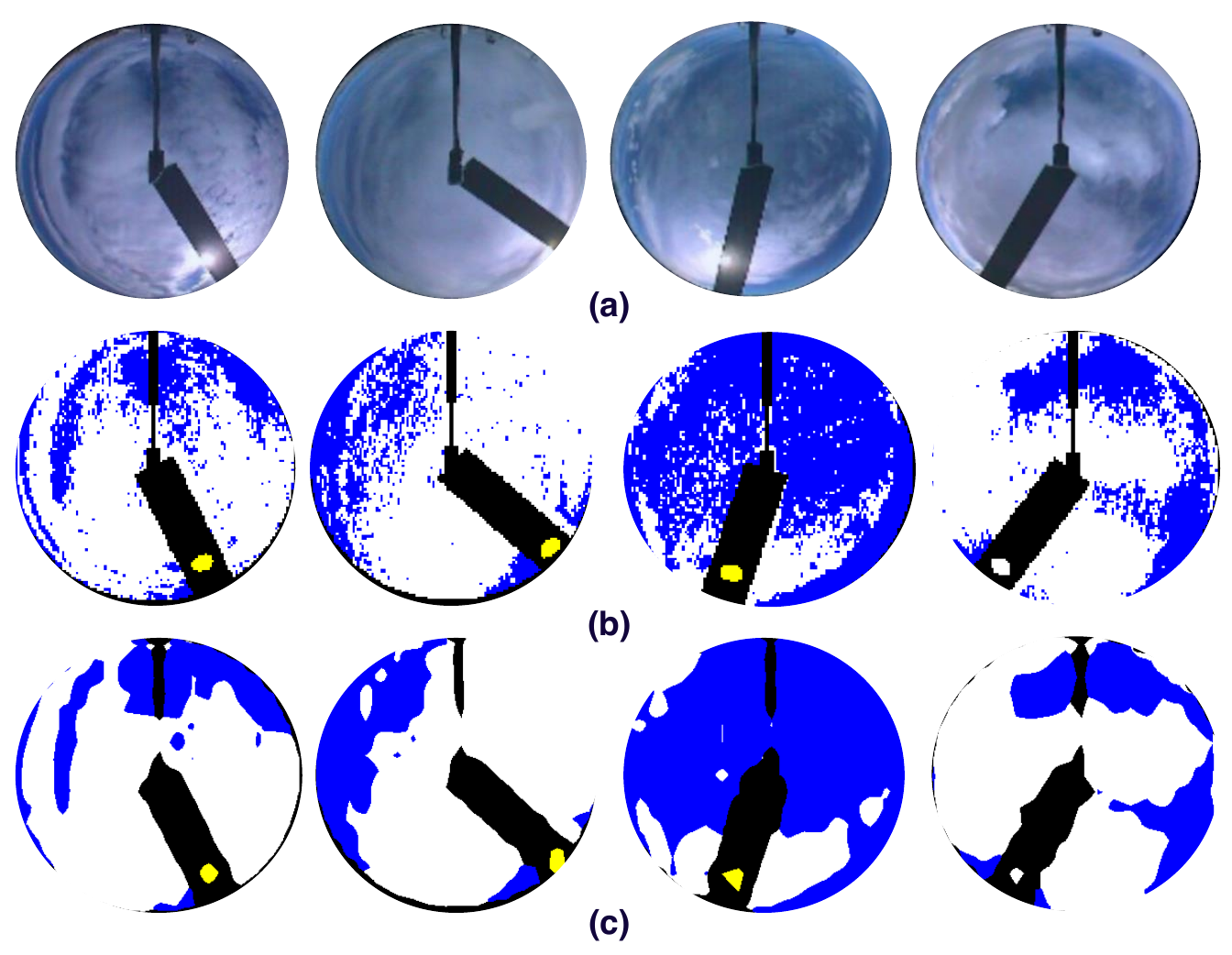}
\caption{\label{fig:attnow} Sample semantic segmentation of \emph{now} predictions. The three rows in the illustration are a sequence of input frames, the corresponding ground truth and semantic masks, respectively.}
\end{figure}

The performance of the \emph{now} prediction model in Table \ref{tab:now} shows that the ResNet50 model (with dilated convolutional filters) captures cloud region with Intersection over Union (IoU) of $83.31\%$ and IoU of $87.89\%$ on sky with the \emph{segment} task on sky-video frames. Simultaneously, the model has a normalized mean absolute error (nMAE) of $10.96\%$ on the \emph{measure} task of solar irradiance. The total accuracy measure which includes the segmentation accuracy of the tracker and boundary bezel of the frame is 94.57\%. As shown in Fig. \ref{fig:attnow}, the ground truth masks are grainy pixel segmentation, whereas our model generates semantic masks by linear up-sampling from a low dimensional representation, resulting in smoother segmentation masks.

Fig. \ref{fig:attnow} also illustrates sample image frames and their corresponding segmentation masks obtained. Unlike rigid objects in benchmark vision datasets, clouds exhibit non-compressible fluid behaviour such as advection and diffusion. The trained \emph{now} prediction model weights are also used as pre-training initialization weights in the \emph{future} prediction experiments. As shown in the \emph{now} experiments in Table \ref{tab:now}, our approach out-performs a direct regression model, trained separately with the loss function corresponding to only \emph{measure}, Eq. \ref{eq:huberloss}. We exclude a comparison of existing literature in skycamera based irradiance measurement, as all deep neural network models far-outperform previous approaches when trained with data from the same location \cite{paoli2010forecasting}. 
The measure task from the \emph{future} prediction model forecasts the solar irradiance for next hour at 10 minutes interval. The performance of the model is improved by spatial attention mechanism. Specifically, the normalized-MAE\% for $10$ minutes forecast of solar irradiance improves from $28.1\%$ without attention, and $21.8\%$ for mean attention, to $19.5\%$ for spatial attention. 

The \emph{segment} task in \emph{future} prediction model forecasts semantic segmentation masks for the next hour at 10 minutes interval. For 10 minutes ahead-of-time prediction, the IoU for cloud segmentation is improved from $61.6\%$ (with no attention) to $74.2\%$ with spatial attention. Similar improvement from $61.2\%$ to $67.7\%$ is observed for 30 minutes ahead-of time predictions. In order to maximize the effect of attention mechanisms, we compute the attention vector only on the $c=cloud$ dimension. We find empirically that this improves the affect of attention in the \emph{future} prediction models. We choose the $cloud$ class, as it is of most value in this application scenario. 

\begin{figure*}[!ht]
\centering
\subfigure
{\includegraphics[width=0.75\textwidth]{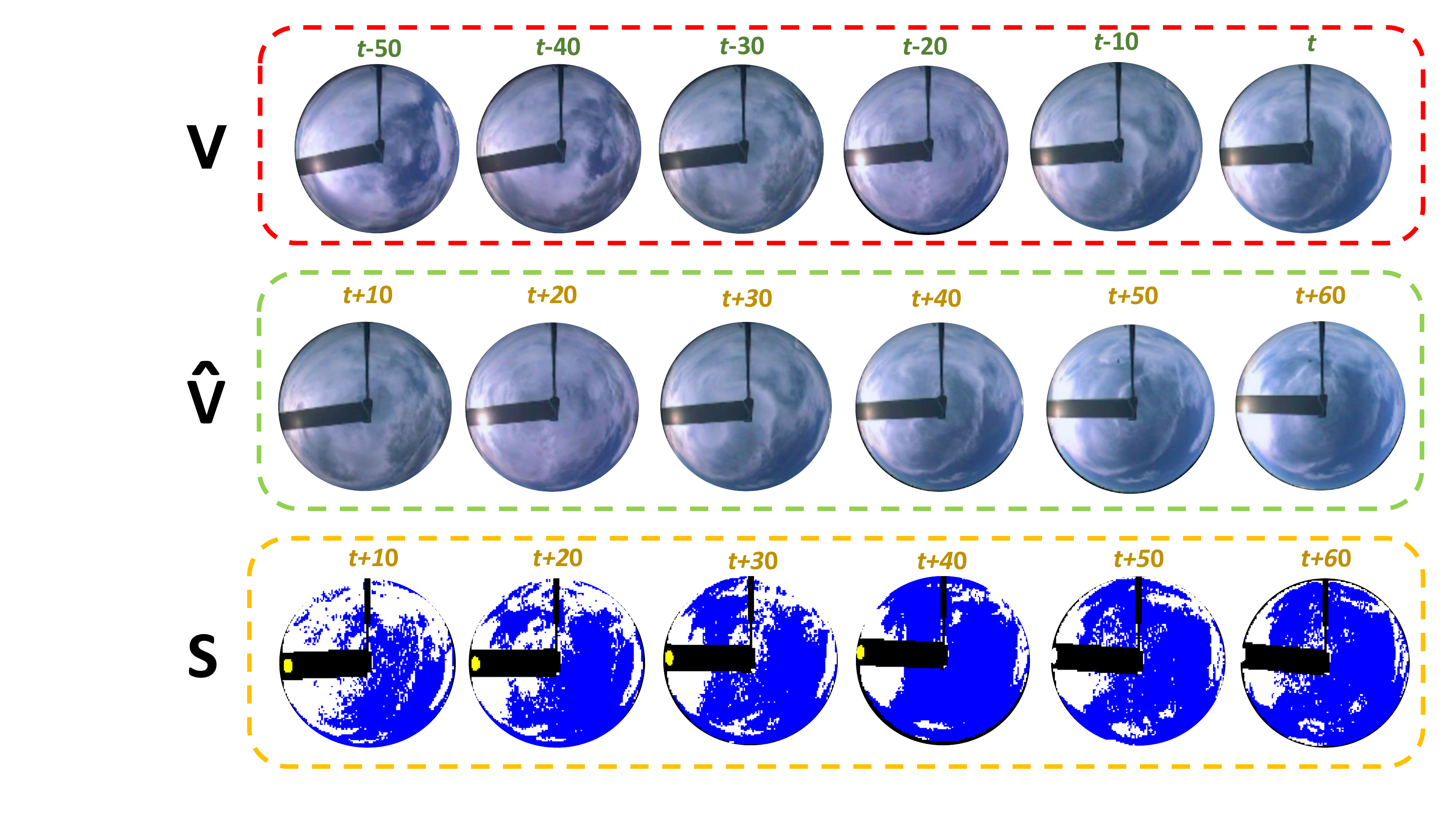}}
\subfigure
{\includegraphics[width=0.75\textwidth]{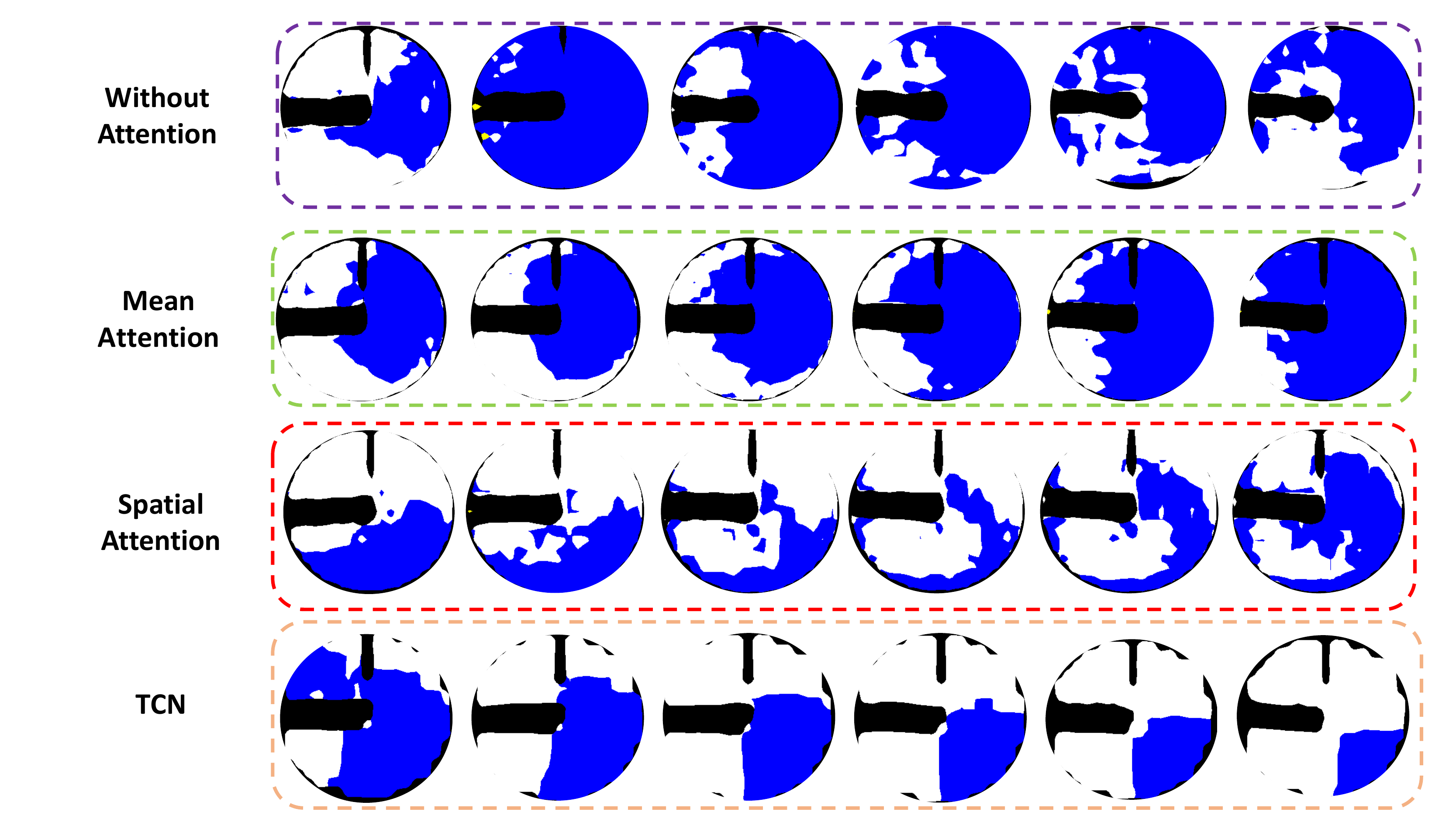}}
\caption{\label{fig:overview} Performance of \emph{segment} task on 6 consecutive frames of one hour. The figure contains input ($V$), target ($\hat{V}$), ground truth ($S$) and 6 predictions (next 1 hour) for multiple attention methods and TCN \cite{neverova2017predicting}.}
\end{figure*}

Fig. \ref{fig:overview}(a) represents a one hour temporal frames ($V$) that is input to the model, the \emph{expected} segmentation masks (${S}$), and the true original future frames($\hat{V}$) for the next one hour prediction. Frames at intervals $t-50$, $t-40$, $t-30$, $t-20$, $t-10$, $t$ are used to generate the semantic segmentation masks for intervals $t+10$, $t+20$, $t+30$, $t+40$, $t+50$ and $t+60$ minutes. Fig. \ref{fig:overview}(b) represents \emph{corresponding} semantic segmentation masks as generated by various attention mechanisms for one hour ahead sky orientation. The first row represents predicted semantic masks without the aid of any attention. Masks predicted for later intervals using this model are not accurate and mis-classify many cloud regions as sky. The next two rows represent semantic masks generated using mean, and spatial attention respectively. It can be inferred that masks generated using spatial attention attend to a more precise representation of sky and hence generate better frames even for the later time intervals.

An empirical analysis of the \emph{partial-}contribution from localized regions of the video frame for the total estimate of solar irradiance requires pixel level ground truth for solar irradiance. However, to our knowledge, such an instrumentation is not available. As discussed above, our architecture design ensures that both \emph{segment} semantic mask of sky region and \emph{measure} of solar irradiance are upsampling estimates from a single $I$ ($fc7$) tensor of course resolution. We assert that the model implicitly localizes the individual contributions of each sky pixel to obtain the total measurement. 

\noindent{\bf Comparison with \cite{neverova2017predicting}}: We use the open-source implementation provided by the authors for this comparison. The base model (pre-trained ResNet with dilated convolution layers) is the same as the proposed model but computed at two scales as per their algorithm \cite{mathieu2015deep}. The summation of $L_1$ loss with gradient difference loss using Adam optimizer. As shown in Table \ref{tab:seg}, the IoU for the first future frame prediction (+10 minutes ahead) is 63.15\%, which is comparable to the persistence model. However, the performance degrades by the third frame (+30 minutes ahead) to 52.65\% and by the sixth frame (+60 minutes) to 39.55\% due to the compounding effects of the auto-regression approach. The bottom row of Fig. \ref{fig:overview}(b) illustrates the effect of compounding errors. 
\section{Conclusion}


We uniquely evaluate our future frame prediction for video understanding on a time-lapse videos of the sky. Our approach outperforms temporal CNNs at future pixel-wise labeling of sky regions. Further, the integral over the spatial regions of the image can  produce an estimate of the total solar irradiance from the sky. The solar irradiance prediction so obtained closely approximate a pyranometer readings over two years test period without re-training or online update, indicating the efficacy of the frame representation. 
Further, the architecture compel the model to learn localized partial-contributions of solar irradiance from different regions of the sky. All scripts will be open-sourced for easy reproducibility (\url{https://bit.ly/2Bw7HGP}).

{
\small
\bibliographystyle{aaai}
\bibliography{skycam}
}

\end{document}


\title{Future frame semantic segmentation of time-lapsed videos with large temporal displacement} 
\author{\bf Supplementary}

\maketitle

\appendix

\noindent {\bf Summary:} We present a novel attention+ConvLSTM based future-frame prediction method with a dual objective over two related tasks on time-lapsed videos: 
\noindent
\emph{Segment}: Semantic segmentation mask prediction for next $n$ future frames.
\noindent
\emph{Measure}: Predict $n$ future values of an additional property of each frame.

\noindent A large publicly available, semantic segmentation dataset is presented with practical application in alternate energy domain. Our approach outperforms state-of-art by 10.8\% for 10 mins (21\% over 60 mins) ahead of time predictions, while maintaining a 10.51\% nMAE on measure task. Our proposed spatial attention model improves the performance by 3.76\% for 10 mins ahead compared to mean attention models.  

The sky-videos are accompanied by two ground truth measurements. The semantic masks are obtained from the TSI imager over a calibrated mirror with a robotic arm. This is a specialized instrument that processes the input feed from the mirror \cite{srrl}. The \emph{measure} task ground truth is obtained from a separate pyranometer sensor. Both measurements are continuously streamed on a public website from Colorado, USA. 

\begin{table*}[!ht] \small
	\begin{center}
		\caption{\label{tab:exp1} Forecasting Architecture with Spatial Attention. \# of Parameter : 27,890,703 }
		\begin{tabular}{|c|c|c|c|}
			\hline \multirow{1}{*}{\textbf{Layer Type}} & \multicolumn{1}{c|}{\textbf{Output shape}} & \multicolumn{1}{c|}{\textbf{\# of Parameters}} & \multicolumn{1}{c|}{\textbf{Connected to}}\\\cline{1-4}
			Input\_Image & (None, 6, 320, 320, 3) & 0 & -\\\hline
			TimeDistributed\_Vision & (None, 6, 20, 20, 4) & 23,595,908 & Input\_Image\\\hline
			Lambda\_Only\_Cloud\_Dimension & (None, 6, 20, 20) & 0 & TimeDistributed\_Vision\\\hline
			Permute  & (None, 20, 20, 6) & 0 & Lambda\_Only\_Cloud\_Dimension \\\hline	
			\textbf{ConvLSTM\_Attention} & (None, 20, 20, 64) & 416,256 & Permute \\\hline	
			BatchNorm\_1 & (None, 20, 20, 64) & 256 & Conv2D\_Attention \\\hline	
			Dense\_Attention & (None, 20, 20, 6) & 390 & BatchNorm\_1\\\hline	
			Reshape & (None, 20, 20, 1, 6) & 0 & Dense\_Attention \\\hline	
			Lambda\_Replicate\_All\_Dimension & (None, 20, 20, 4, 6) & 0 & Reshape \\\hline	
			Permute\_Attention\_Vector & (None, 6, 20, 20, 4) & 0 & (None, 6, 20, 20, 4) \\\hline	
			Attention\_Multiply & (None, 6, 20, 20, 4) & 0 & TimeDistributed\_Vision \textit{*} Permute\_Attention\_Vector\\\hline	
			ConvLSTM\_1 & (None, 6, 20, 20, 128) & 1,690,112 & Attention\_Multiply \\\hline
			BatchNorm\_2 & (None, 6, 20, 20, 128) & 512 & ConvLSTM\_1 \\\hline
			ConvLSTM\_2 & (None, 6, 20, 20, 64) & 1,229,056 & BatchNorm\_2 \\\hline
			BatchNorm\_3 & (None, 6, 20, 20, 64) & 256 & ConvLSTM\_2 \\\hline
			ConvLSTM\_3 & (None, 6, 20, 20, 64) & 819,456 & BatchNorm\_3 \\\hline
			BatchNorm\_4 & (None, 6, 20, 20, 64) & 256 & ConvLSTM\_3\\\hline	
			Conv3D(f.c. layer) & (None, 6, 20, 20, 4) & 6,404 & BatchNorm\_4 \\\hline	
			\textbf{Segment (Bilinear Upsampling)} & (None, 6, 320, 320, 4) & 0 & Conv3D \\\hline
			Conv2D\_1 & (None, 6, 18, 18, 128) & 4,736 & Conv3D \\\hline 
			BatchNorm\_5 & (None, 6, 18, 18, 128) & 512 & Conv2D\_1 \\\hline
			Activation\_ReLu\_1 & (None, 6, 18, 18, 128) & 0 & BatchNorm\_5 \\\hline
			Conv2D\_2 & (None, 6, 16, 16, 128) & 147,584 & BatchNorm\_5 \\\hline
			BatchNorm\_6 & (None, 6, 16, 16, 128) & 512 & Conv2D\_2 \\\hline
			Activation\_ReLu\_2 & (None, 6, 16, 16, 128) & 0 & BatchNorm\_6 \\\hline
			Flatten & (None, 6, 32768) & 0 & Activation\_ReLu \\\hline
			Dropout (0.5) & (None, 6, 32768) & 0 & Flatten \\\hline
			\textbf{Dense\_Measure} & (None, 6, 1) & 32,769 & Dropout (0.5) \\\hline	
		\end{tabular}
	\end{center}
	\begin{center}
		\caption{\label{tab:exp1} Spatial Attention (conv2D) pipline. Reduces \# of Parameters of the entire model to 27,484,111 }
		\begin{tabular}{|c|c|c|c|}
			\hline \multirow{1}{*}{\textbf{Layer Type}} & \multicolumn{1}{c|}{\textbf{Output shape}} & \multicolumn{1}{c|}{\textbf{\# of Parameters}} & \multicolumn{1}{c|}{\textbf{Connected to}}\\\cline{1-4}
			\textbf{Conv2D\_Attention} & (None, 20, 20, 64) & 9,664 & Permute \\\hline	
			BatchNorm\_1 & (None, 20, 20, 64) & 256 & Conv2D\_Attention \\\hline	
			Dense\_Attention & (None, 20, 20, 6) & 390 & BatchNorm\_1\\\hline	
			Reshape & (None, 20, 20, 1, 6) & 0 & Dense\_Attention \\\hline	
			Lambda\_Replicate\_All\_Dimension & (None, 20, 20, 4, 6) & 0 & Reshape \\\hline	
			Permute\_Attention\_Vector & (None, 6, 20, 20, 4) & 0 & (None, 6, 20, 20, 4) \\\hline	
			Attention\_Multiply & (None, 6, 20, 20, 4) & 0 & TimeDistributed\_Vision \textit{*} Permute\_Attention\_Vector\\\hline	
		\end{tabular}
	\end{center}
\end{table*}

\section{Model architectural details}
The model architecture details are presented here:
\begin{itemize}
\item
The RGB input image of size $(320,320,3)$ is first passed through a ResNet50 Architecture \cite{yu2015multi} with kernel regulariser set as \textit{$l_2$-norm}, weight decay of 5e-5, and batch momentum of 0.9 throughout.

\begin{figure}[!b] 
\centering
\subfigure[\small{bloom}]{\includegraphics[width=0.2\textwidth]{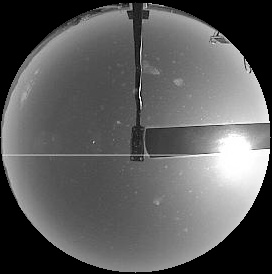}} 
\subfigure[\small{track}]{\includegraphics[width=0.2\textwidth]{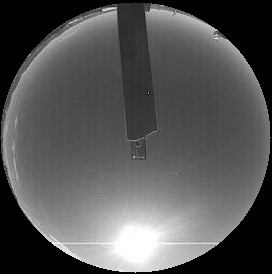}} 
\subfigure[\small{rain}]{\includegraphics[width=0.2\textwidth]{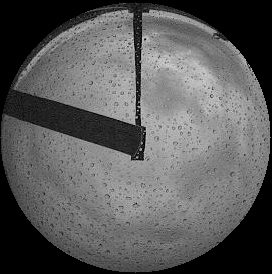}} 
\subfigure[\small{dew}]{\includegraphics[width=0.2\textwidth]{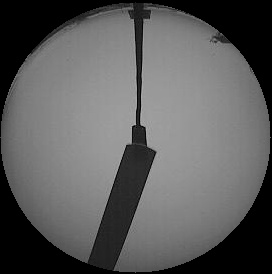}}
\caption{\label{fig:chal} Challenging sky-video frames from the dataset. (a) Blooming effect of bright sun over the occluder, (b) miss-aligned occluder, (c) heavy rain and droplets, (d) dew on the lens after sunrise.}
\end{figure}
	
\item
	The resulting vector is then forwarded to an Atrous block with output size $(20,20,4)$.This is referred to as the representation vector $I$ in the Architecture Section. We use bilinear upsampling to produce semantic segmentation of $(320,320,4)$ corresponding to the $4$ classes.
	\item
	For $measure$, the same output from the Atrous block is passed through 2 blocks of Conv2D layer with 128 filters of kernel size $(3,3)$, kernel regulariser set as \textit{$l_2$-norm} and weight decay of 5e-5 followed by BatchNorm and ReLU activation. The resultant vector of size $(16,16,128)$ is flattened and densely connected to 1 unit $measure$ prediction output with $0.5$ Dropout. The intuition is explained in Model Section.
	
	\item
	For forecasting, all experiments are performed with a look back ($t$) and forward predict of 6 frames (corresponding to 1 hour). For each consecutive video frames set, 6 representation vectors ($\mathbb{I}$) corresponding to the 6 input frames are obtained. 
	
	\item
	The three-tier ConvLSTM architecture is used to predict the representations of the next 6 frames of size $(6,20,20,4)$. The forecasting architecture consists of three blocks of ConvLSTM2D with 128, 64, and 64 filters respectively each with filter size $(5,5)$. The ConvLSTM2D layers are alternated with BatchNorm and the output is reshaped to $(6,20,20,4)$ using a Conv3D layer with 4 filters of kernel size $(1,5,5)$. The remaining attention vectors are also scaled similarly. 
	
	\item
	The forecasted representation vector $I$ for each time step can be resolved into \textit{segment} and \textit{measure} as described above.
	
	\item
	The performance of the three-tier ConvLSTM is improved with attention multipliers over the input representation set ($\mathbb{I}$). The multipliers spatially and temporally (element-wise) weights the input vector, and hence is of the same size. 
	
	\item
	To produce an attention multiplier of size $(6,20,20,4)$, we first select only the cloud segment of size $(6,20,20,1)$. For spatial attention multiplier, Conv2D of 64 filters with size $(5,5)$ is applied with BatchNorm over the \textit{temporal} dimension. It is then densely down-sampled to number of samples (6) with $softmax$. This provides a bounded attention value (0-1) over the look back period (6) constrained spatially.
	
	
	\item
	The attention obtained from the cloud segment channel is replicated to all segmentation classes. Figure 2 illustrates a sample attention mask over laid on the original frames obtained from the spatial attention model described above. 
	
	\item
	While we perform experiments on several variations of this architecture in the empirical evaluation, these experiments on architecture search are not exhaustive. However, it is our understanding that the model sufficiently out performs several satellite based meteorological techniques for solar energy prediction currently in use today.

	\item 
	Table 3 lists some of the salient properties of the sky-video dataset. The sky-video dataset is larger and has a wider temporal gap as compared to other publicly available datasets. Some challenging frames from the dataset are illustrated in Fig. \ref{fig:chal}.
\end{itemize}

\begin{figure*}[!ht]
\begin{center}
\includegraphics[width=\textwidth]{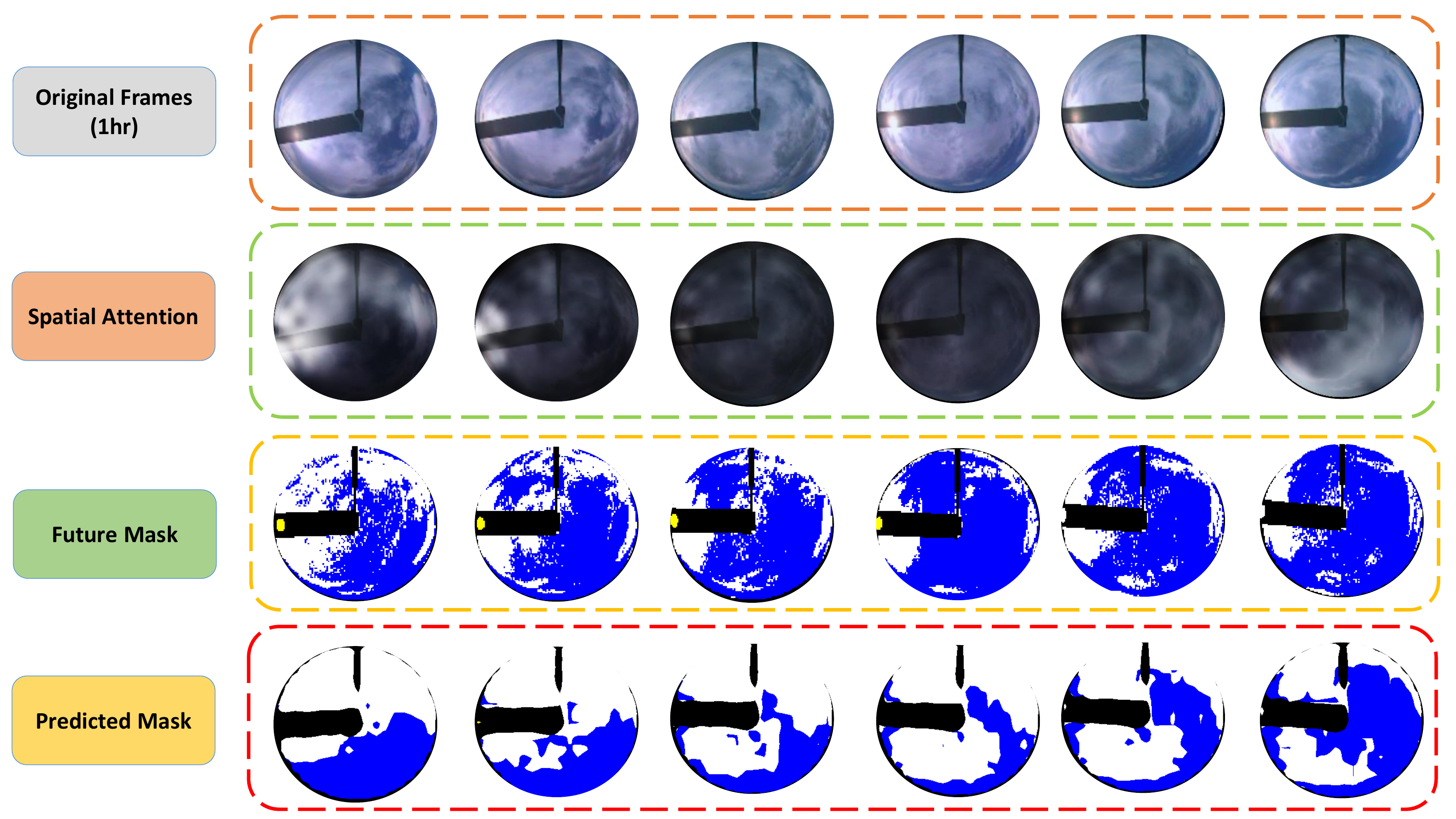}
\end{center}
\caption{\label{fig:overview}: Performance of \emph{segment} task on a challenging set of six consecutive sequence of frames representing one hour. The illustration contains input, attention-overlay, target, and predicted sequences.}
\end{figure*}


\begin{table*}[!ht] 
\newlength\qq
\setlength\qq{\dimexpr .3\textwidth}
\centering
\caption{Salient properties of Sky-video dataset}
\label{tab:dbcomp}
\noindent\begin{tabular}{p{\qq} p{\qq} p{\qq}}
\hline
{\bf Properties} & {\bf Sky-video} \\ \hline
\# of train/val  & 73,120/49,944  \\ \hline
total size & 123,064 \\ \hline
frame rate  & 6 frames/hour \\ \hline
ground truth  & every frame \\ \hline
\# of classes  & 4 \\ \hline
\end{tabular}
\end{table*}

\section{Related work in weather prediction}
The earliest methods for weather prediction were geometry based models with strong assumptions on functional dependency of position, time and location \cite{hor}. Highly accurate weather prediction models that are based on coarse grained simulation of physical weather systems are predominantly used in weather prediction \cite{ecmwf,gfs}. However, such complex systems have a systematic bias to certain location, time, weather phenomenon, or unpredictable weather occurrences. Recently, several ensemble based techniques have been introduced, that combine physics models with numerical and purely data driven models to fine-tune predictions \cite{lu15}. While such approaches show substantial improvement in forecast accuracy, they are limited by the availability of satellite data (typical satellite sweeps range between 3-12hrs), need for enormous computational infrastructure and inability to perform short-term corrections to predictions. Certain weather parameters can be predicted in short-term horizons with suitable local sensor deployment in a region. Solar-irradiance is one such measure that can be sensed with varying degrees of accuracies. Achleitner   \cite{achleitner2014sips} present a approach to aggregate several small photo-sensors for predicting irradiance. While Aryaputera   \cite{Aryaputera20151266} present a regression approach to extrapolate weather information to unknown locations. Su   \cite{su2015local} present a local feature approach to explicitly segment and track each cloud with a  adaptive gaussian mixture model approach, followed by hand-crafted features for matching clouds across frames for tracking. Other pixel clustering and segmentation based approaches  \cite{wacker2015cloud,heinle2010automatic} explicitly measure the cloud cover from sky-images in terms of meteorological unit of \emph{okta} (the number of eighths of the sky occluded by clouds) and cloud type.


{\small
\bibliographystyle{aaai}
\bibliography{skycam_ap}
}